\documentclass[letterpaper, 10 pt, conference]{ieeeconf}
\IEEEoverridecommandlockouts
\usepackage{xcolor}  
\usepackage{booktabs}
\usepackage{fancyhdr}
\usepackage{graphics}
\usepackage{graphicx}
\usepackage{epsfig}
\usepackage{color}
\usepackage[format=plain,font=small,labelfont=it,labelsep=period]{caption}
\usepackage{subcaption}
\usepackage{wrapfig}
\usepackage{float}
\usepackage{cite}
\usepackage{times}
\usepackage{amsmath}
\usepackage{amssymb}
\usepackage{amsfonts}
\usepackage{mathtools}
\usepackage{url}
\usepackage{algorithm}
\usepackage{lipsum}
\usepackage{algpseudocode}
\usepackage{hyperref}

\title{\LARGE \bf
Vision-Guided Quadrupedal Locomotion in the Wild with \\ Multi-Modal Delay Randomization
}

\author{
Chieko Sarah Imai\textsuperscript{*1} \quad 
Minghao Zhang\textsuperscript{*2} \quad 
Yuchen Zhang\textsuperscript{*1} \quad 
Marcin Kierebiński\textsuperscript{1} \quad
Ruihan Yang\textsuperscript{1} \\
Yuzhe Qin\textsuperscript{1} \quad
Xiaolong Wang\textsuperscript{1} 
\thanks{\textsuperscript{1} University of California, San Diego, CA, USA }% <-this
\thanks{\textsuperscript{2} Tsinghua University, Beijing, China}
\thanks{* Equal contributions. Names ordered alphabetically.}
}

\begin{document}

\makeatletter
\let\@oldmaketitle\@maketitle% Store \@maketitle
\renewcommand{\@maketitle}{\@oldmaketitle% Update \@maketitle to insert...
  \includegraphics[width=\textwidth]{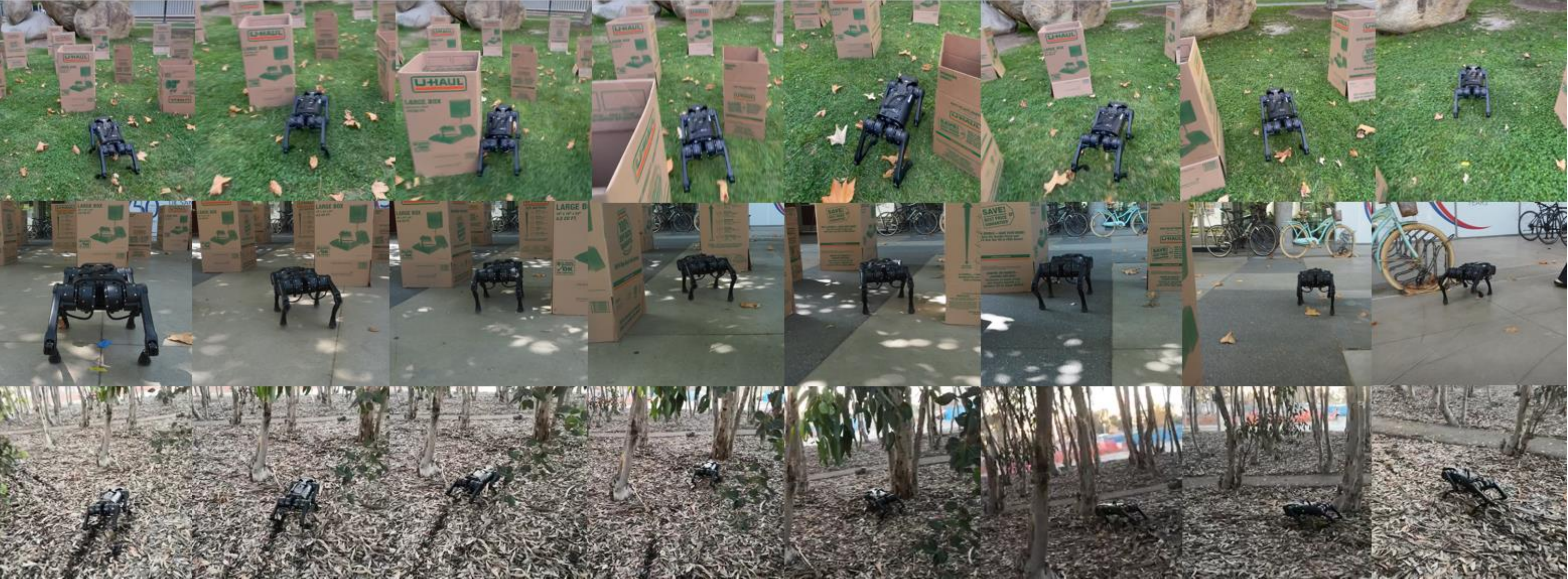}
  \vspace{-0.2in}
  \captionof{figure}{Our method enables quadruped robots to traverse complex environments with obstacles of different shapes in the wild. The learned locomotion policy obtains different agile locomotion skills like flexibly turning before barriers and stably walking on steep lawn, using reinforcement learning. The policy is trained in simulation with our multi-modal delay randomization and deployed in the real-world, unseen scenarios without any adaptation or fine-tuning.
  }
    \label{fig:teaser}
    \vspace{-0.17in}
}

\newcommand{\revision}[1]{\textcolor{blue}{#1}}

\thispagestyle{empty}
\pagestyle{empty}
\maketitle 
\setcounter{figure}{1}

\begin{abstract}

Developing robust vision-guided controllers for quadrupedal robots in complex environments with various obstacles, dynamical surroundings and uneven terrains is very challenging. While Reinforcement Learning (RL) provides a promising paradigm for agile locomotion skills with vision inputs in simulation, it is still very challenging to deploy the vision-guided RL policy in the real world. Our key insight is that the asynchronous multi-modal observations, caused by different latencies in different components of the real robot, create a large sim2real gap for a RL policy.  In this paper, we propose \emph{Multi-Modal Delay Randomization (MMDR)} to address this issue when training in simulation. Specifically, we randomize the selections for both the proprioceptive states and the visual observations in time during training, aiming to simulate the asynchronous inputs when deploying to the real robot. With this technique, we are able to train a RL policy for end-to-end locomotion control in simulation, which can be directly deployed on the real A1 quadruped robot running in the wild. We evaluate our method in different outdoor environments with complex terrain and obstacles. We show that the robot can smoothly maneuver at a high speed while avoiding the obstacles, achieving significant improvement over the baselines. Our project page with videos is at {\color{blue} \href{https://mehooz.github.io/mmdr-wild/}{https://mehooz.github.io/mmdr-wild/}}.

\end{abstract}
\section{Introduction}

Learning legged locomotion is one of the most challenging problems in robotics. The goal is to allow the robot to traverse complex, uneven, and in-the-wild environments~\cite{mitcheetah2018mpc, di2018dynamic}.
Reinforcement Learning (RL) has brought significant improvements in legged locomotion~\cite{motion_imitation, hwangbo2019learning, Da2020LearningAC}, where the policies are trained with proprioceptive state inputs. Specifically, we have witnessed impressive control performance in the wild using RL incorporating robustness and adaptive ability~\cite{lee2020learning, zhaoming2020DR, kumarFPM21}.

RL policies with only proprioceptive information in the observation space have shown promising results on maneuvering on uneven and unknown-material grounds. Yet this is insufficient for more complicated tasks, such as avoiding obstacles that are hard to step over, or estimating accurate foot placement positions. To overcome these challenges, the robot needs to perceive the obstacles and terrains ahead to plan the path to move. In this work, we propose to train RL end-to-end, with both the proprioceptive states and depth image from the camera on the front of the robot's head as inputs, that outputs the robot joint control action. Our hypothesis is that the proprioceptive states can provide information about the immediate or prior interaction with the terrain, and visual observations provide information about the near-future terrain for better planning. With both types of information, it allows our robot to navigate in challenging, in-the-wild scenarios as Figure~\ref{fig:teaser} demonstrates.

However, learning with multi-modal inputs with both visual observations and proprioceptive states introduces new challenges in Sim2Real transfer for RL. While in simulation every signal is aligned, in the real robot, the sensors' transmission, the control algorithm's computation, and the robot's response all introduce latency for perception and control. The problem is further magnified when control requires both high-dimensional vision and proprioceptive state inputs. This introduces two challenges to RL. First, the deficiency of on-board computing resources in a quadrupedal robot causes  latency in processing the visual inputs, which will delay the action output from the RL policy causing danger to the robot when running in high speed. Thus the policy will need to be robust to the visual delay and still perform correct decision making. Second, since the latency caused by different sensors are different, it introduces asynchrony among multi-modal inputs for the RL policy: there is misalignment between the proprioceptive state and the visual observation in the real robot. While this latency and misalignment of information can be modeled in classical control pipelines~\cite{fubinf-b08-03}, \textbf{how to address the asynchronous multi-modal inputs for RL policies still remains an open problem}. Solving the two challenges above is essential for transferring a vision-guided RL policy trained in simulation to the real robot.

In this paper, we propose \emph{Multi-Modal Delay Randomization (MMDR)} to improve the robustness of multi-modal RL policies in real robots. The key is to simulate the asynchrony in the real world during training. During learning in simulation, instead of forwarding both proprioceptive states and visual observations with fixed temporal intervals, we randomize the inputs to make them asynchronous for training the policy network. 
Specifically, we maintain two buffers online for two types of inputs. Each buffer will store one type of recent stream observations. We independently sample a sub-sequence of observations from each buffer, and forward both types of inputs to the policy network. The multi-modal policy trained with RL using MMDR is robust to the delayed visual inputs and the misalignment between two types of observations for locomotion control.

We experiment with our proposed method on challenging, in-the-wild maneuvering tasks with a real A1 quadruped robot~\cite{a12021} as shown in Figure~\ref{fig:teaser}. These environments include static or moving obstacles of different sizes and shapes, uneven terrains, and changing lighting conditions. We show that applying MMDR during training significantly improves the performance and robustness for real-world deployment of the learned RL policy. Our policy can even generalize to environments with unseen or moving obstacles, where the A1 robot plans and maneuvers smoothly through the obstacles at a high moving speed. 

\section{Related Work}

\noindent\textbf{\emph{Learning-based Legged Locomotion.}}
Controlling a legged robot has been studied by the robotics community for a long time. Control theory and trajectory optimization approaches have shown great results on legged locomotion control~\cite{mitcheetah2018mpc,di2018dynamic,trajectoryopt2019,ding2019real}, however these methods require in-depth knowledge about the environment or substantial manual efforts for parameter tuning. As an alternative, model-free Reinforcement Learning provides a generalizable learning paradigm for legged locomotion skills from self-exploration in complex environments~\cite{motion_imitation,tan2018sim2real, hwangbo2019learning,lee2020learning,peng2017deeploco, zhaoming2020DR,pmtg}. Despite the successful application of RL on legged robots, most approaches are still training a ``blind'' robot with only proprioceptive inputs. This limits the robot from maneuvering over challenging environments where large obstacles are presented. Previous works have also considered using a high-level visual planner to guide the locomotion controller~\cite{hoeller2021learning,fu2021coupling, li2021vision} for navigation. Instead of training policies in a hierarchical manner, we provide an end-to-end RL framework which takes vision and proprioceptive states as inputs and directly performs the robot joint control. This simple yet effective approach allows the robot to robustly navigate challenging terrains in the wild.

\noindent\textbf{\emph{Reinforcement Learning with Visual Inputs.}}
To enrich the perception of the agent, many works have studied RL with visual input in  navigation~\cite{midLevelReps1,FaustORFTFD18,DBLP:conf/iclr/WijmansKMLEPSB20}, continuous control~\cite{rad,hansen2021softda,drq,hansen2021stabilizing}, and manipulation~\cite{levine-e2e,levine-eye-co}. Inspired by these works, recent works also introduce to perform RL with visual observation in end-to-end~\cite{yang2021learning} or hierarchical manners~\cite{jain2020pixels} for legged robot locomotion control. Multi-modal inputs including vision are also leveraged for locomotion control~\cite{deepmind-loco-tog, margolis2021learning}. However, most of the previous works are still focusing on running the robot in simulation or in a lab environment. In this paper, we show that our approach enables the vision-guided quadruped robot to maneuver at a high speed on challenging terrains, avoiding large obstacles in the wild.

\noindent\textbf{\emph{Reinforcement Learning with Domain Randomization.}}
To address the sim2real gap for RL policy deployment in the real-world, domain randomization introduces variability to different components of the simulation environment during training, including  variant physical parameters~\cite{DBLP:conf/iros/MordatchLT15, PengAZA18,tan2018sim2real, li2021reinforcement}, visual attributes~\cite{tobin2017domain, DBLP:journals/ijrr/OpenAI20}, and force perturbations~\cite{zhaoming2020DR, DBLP:journals/ijrr/OpenAI20}. Due to its simplicity and effectiveness, domain randomization has been widely used in learning legged robot policy to complete real-world tasks with diverse locomotion skills~\cite{xie2020learning, motion_imitation, kumarFPM21, lee2020learning}. However, none of the previous work has addressed the problem of asynchronous multi-modal inputs for running RL policy on real legged-robots. To address this problem, we propose MMDR during training RL policy in simulation, approximating the asynchronous sensory signals in the real world.

\noindent\textbf{\emph{Learning with Delay.}}
The perception and control latency recently draws attention from both the vision and the RL community. For example, in computer vision, Li et al.~\cite{Li2020StreamingP} introduces streaming perception to consider the latency of the system when performing visual recognition in the real world. In Reinforcement Learning, continuous-time RL~\cite{continuous-RL} and MDP with delay~\cite{mlp-delay} has been proposed to address constant system latency in simulation. Beyond simulation, the continuous-time algorithms are developed for concurrent control~\cite{Xiao2020Thinking, Lutteretal21b} in real-world control tasks. Our work is inspired by these previous works, and our focus is on Reinforcement Learning with asynchronous multi-modal inputs for transferring to real legged robots, which has rarely been addressed before. 

\section{Reinforcement Learning for Multi-Modal Locomotion}
Our method considers both proprioceptive states and visual observations for locomotion control. We use deep reinforcement learning (DRL) to learn the multi-modal locomotion policy. In RL, at each time step $t$, the agent provide action $a_{t}$ based on the observation $O_{t} = (o^{vis}_t, o^{prop}_t)$ (Here our observation consist of both proprioceptive state $o^{prop}_tt$ and visual observations $o^{vis}_t$), and receive reward $r_t$ as the feedback from the environment. The agent learns a $\theta$-parameterized policy $a_t = \pi_\theta (o^{vis}_t, o^{prop}_t)$ which maximizes the discounted summed reward,
$\mathbb{E}_{\tau\sim p_\theta(\tau)} [ \sum\limits_{t=0}^H \gamma^t r_t]$,
where $\tau$ is the trajectory distribution, the H is the horizon of the environment, and the $\gamma$ is discounted factor. 

\subsection{Network Architecture with Multi-Modal Inputs}
\begin{figure}[t]
\centering
\includegraphics[width=0.9\linewidth]{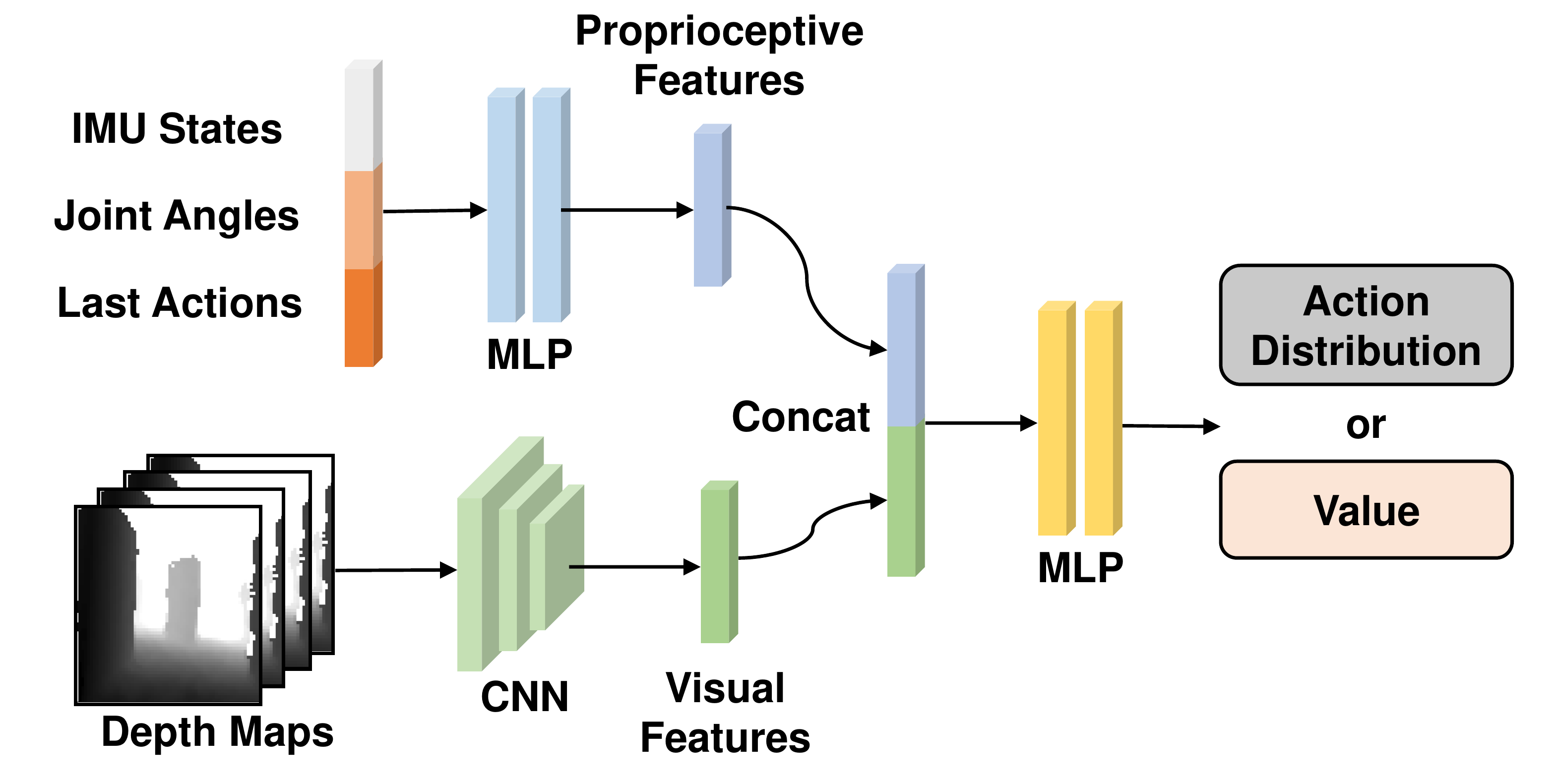}
\vspace{-0.1in}
\caption{\emph{Network Architecture.} We use separate encoders for multi-modal inputs to get domain-specific features and use a MLP to get the value or action distribution from concatenated features.}
\vspace{-0.2in}
\label{fig:network}
\end{figure}

To utilize multi-modal information, both our policy network and value network use the architecture as shown in Figure~\ref{fig:network}. We use a 2-Layer MLP encoder for proprioceptive input and a 3-Layer CNN for visual input to get encoded features from both modalities. We then concatenate the encoded features to get a unified feature and feed it into the additional 2-Layer MLP to predict the action distribution (for policy network) or the value (for value network). For better sample efficiency and stable training, the CNN for visual input is shared across the policy and the value function. 

\noindent\textbf{\emph{Observation Space.}}
The network inputs contain both the proprioceptive state and visual observation. The proprioceptive state consists of (i) 12D robot joint rotation, (ii) 4D IMU sensor reading (angle and angular velocity of roll and pitch), (iii) 12D last executed action. The proprioceptive input (84D) contains three proprioceptive states. The visual input consists of four stacked depth images of shape $64\!\times\! 64$. All the depth images come from the depth camera mounted on the head of the robot. To constrain the scale of the visual observation, depth values are clipped to $[0.3, 10]$m. Note we only use depth images without RGB channels for visual observation. The geometric structures of the environment can be more easily captured by depth than the RGB channels. With RGB channels, it introduces more challenges for Sim2Real transfer, since  we will need to consider the lighting and texture during rendering to match the real world.

\noindent\textbf{\emph{Action Space.}}
Our action space for the policy is the 12-D target joint angle for each joint of the robot. Target angles are converted to motor torques using a default PD controller. This can be automatically mapped from joint angles to the steering angle for avoiding obstacles and path planning.

\noindent\textbf{\emph{Rewards.}} 
In general, our reward function encourages the robot to safely move forward (along the $x$-axis in simulation) with a target speed (0.35m/s) while minimizing energy cost. 
Specifically, the reward function is given by:

\emph{a) Moving forward reward} encourages the robot to move forward in the target velocity, given by:
$R_{\textnormal{forward}} = \max(\min(0, v),v_{\textnormal{target}})$,
where $v$ is the body velocity along the $x$-axis in world frame, and $v_{\textnormal{target}}$ is the target velocity.

\emph{b) Alive reward.} This term gives $R_{\textnormal{alive}}=1.0$ at each time step until termination. This term is similar to ''Collision Penalty'' in previous works, but won't discourage the robot too much and make it unable to move in the early stage.

\emph{c) Energy reward} encourage the robot to use minimal energy. We penalize the motor torques for all the joints $\tau$ as $R_{\textnormal{energy}}=-\|\tau\|^2$.

the final reward is the weighted sum of all mentioned reward terms, with the weights as follow:
\begin{equation}
\label{rew:total}
\begin{aligned}
R &= R_{\textnormal{forward}} +0.1R_{\textnormal{alive}}+0.005 R_{\textnormal{energy}}.
    \end{aligned}
\end{equation}

\noindent\textbf{\emph{{Training Details.}}}
We use Proximal Policy Optimization (PPO)~\cite{ppo} to train all methods with 10M samples. The encoded features from both modalities are 256D vectors. We use a batch-size of 16384 and split it into 16 mini-batches. We use Adam optimizer~\cite{DBLP:journals/corr/KingmaB14} with learning rate of 1e-4 for both policy and value networks.

\section{Multi-Modal Delay Randomization for Sim2Real Transfer}

Reinforcement Learning with multi-modal inputs introduces extra challenges for Sim2Real transfer. We will first introduce the latency problem in the real-robot system, and how it affects the RL policy. Then we propose a novel Multi-Modal Delay Randomization (MMDR) approach to improve Sim2Real transfer.

\subsection{Asynchronous Multi-Modal Inputs caused by Latency}

\label{sec:Latency}
\begin{figure}[t]
\vspace{0.05in}
\centering
\includegraphics[width=0.9\linewidth]{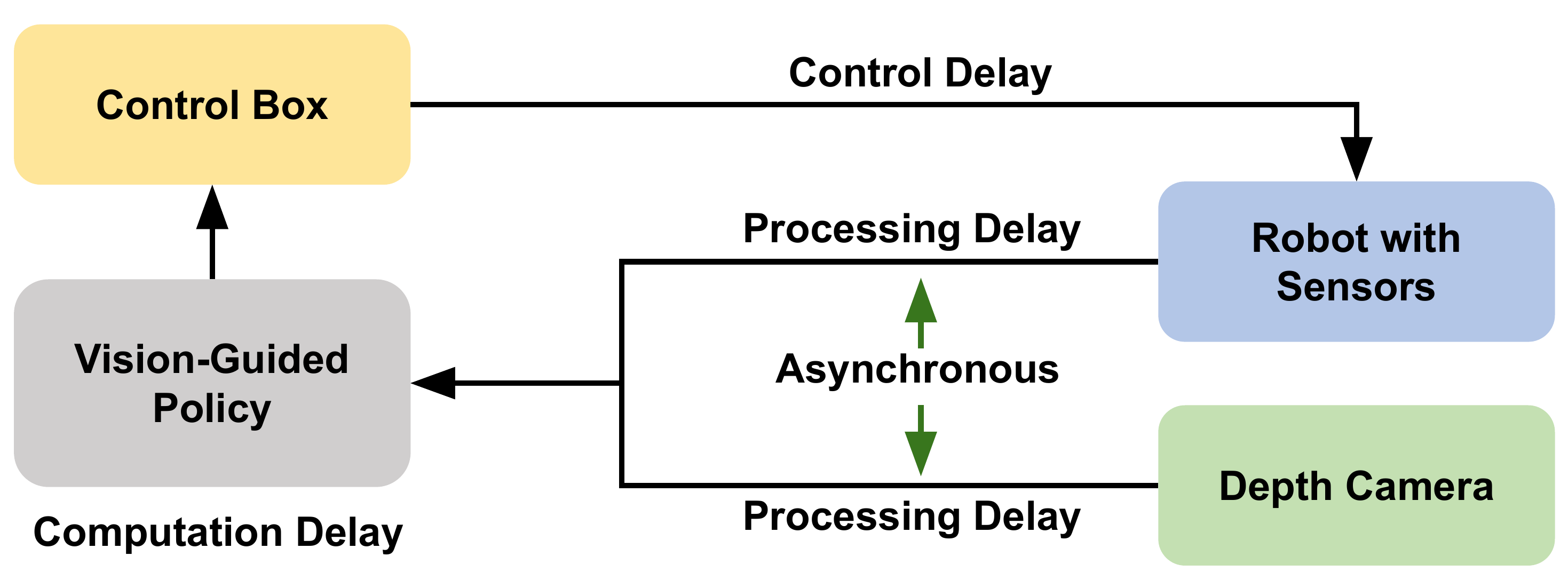}
\vspace{-0.05in}
\caption{\emph{Multi-modal Control System and Delays.} We consider delays in perception procession, policy computation, and control execution.}
\vspace{-0.15in}
\label{fig:delay_demostration}
\end{figure}

When executing the learned multi-modal policy in the real-world, the robot receives perception from different sensors, then the policy computes the control signal, and finally the control signal is applied to the robot. Latency exists in every stage of the multi-modal control system. As illustrated in Figure~\ref{fig:delay_demostration}, the multi-modal control system mainly has three kinds of latency: (i) Processing Latency: time for sensor reading and data processing; (ii) Computation Latency: time for control signal computing; (iii) Control Latency: the time for the control signal to take effect on the motors of the robot. In multi-modal control systems, due to the limited on-board computation resources and the high-dimensional visual input, the computation latency is magnified. By the time the control signals are applied, the surroundings of the robot has already changed. Because of the different mechanism of different sensors, the multi-modal observations are asynchronous when forwarding to the RL policy. However, in common physics simulators~\cite{bullet, mujoco,gazebo, isaac}, the simulated world is considered fixed with perfectly aligned observations. This discrepancy between simulation and the real robot can cause large performance drop when applying Sim2Real transfer for the RL policy.

\subsection{Case Study for Real Robot with Multi-Modal Inputs}
We perform a real-robot case study to obtain a better understanding of sensory latency and how it affects RL. Table~\ref{table:benchmark_sensor_latency} and~\ref{table:benchmark_network_inference_latency} shows the latency of each part in the real robot control pipeline. We observe the latency between the depth map sensor and proprioceptive state senor is heavily asynchronous in the real robot. When computing the inference time of the RL policy networks, we also find there is much larger latency when taking both vision and state as inputs, comparing to network with only proprioceptive state as inputs. Both observations introduce great challenges to transferring an RL policy to the real robot. 

\begin{table}[t]
\centering
\begin{small}
\begin{tabular}{c|c}
\toprule
Module & Module Frequency Inverse $\pm$ Latency (s)  \\
\midrule
Depth Map Sensor & $0.033\pm$ 0.004\\
Joint State Sensor & $0.0025\pm 0.001$ \\
IMU Sensor & $0.0025 \pm 0.001$\\
Network Inference &  $ 0.040 \pm 0.009$\\
Actuation Time& $0.0025 \pm 0.001$\\ 
\bottomrule
\end{tabular}
\vspace{-0.05in}
\caption{Latency (standard deviation) of each part in the real robot. Note, the joint state sensor, IMU sensor and actuation time are all part of the same process interfacing with the robot SDK. Hence the sensor frequency and latency are identical for these sensors.}
\label{table:benchmark_sensor_latency}
\end{small}
\vspace{-0.05in}
\end{table}
\begin{table}[t]
\centering
\begin{small}
\begin{tabular}{c|c}
\toprule
Network Inference Time & Time Taken $\pm$ Latency (s)  \\
\midrule
Network Inference (State Only)& $ 0.004 \pm 0.026 $\\
Network Inference (State\&Vision) & $ 0.040 \pm 0.007 $\\
\bottomrule
\end{tabular}
\vspace{-0.05in}
\caption{Network inference time and the latency observed.}
\label{table:benchmark_network_inference_latency}
\end{small}
\vspace{-0.2in}
\end{table}

\subsection{Multi-Modal Delay Randomization}

\begin{figure}[t]
\centering
\vspace{-0.03in}
\includegraphics[width=0.9\linewidth]{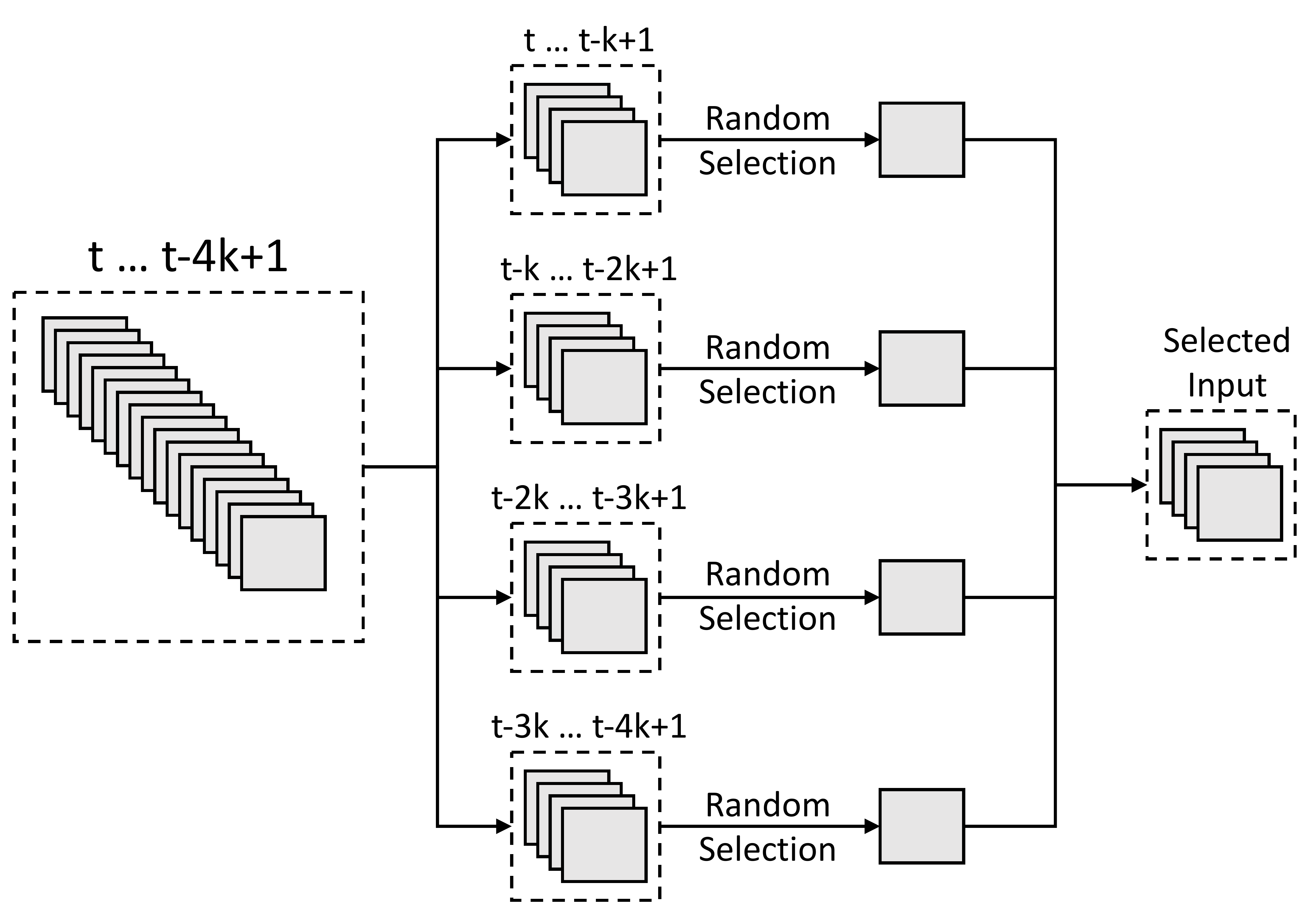}
\vspace{-0.075in}
\caption{\emph{Visual Delay Randomization.} We maintain a depth map buffer with a length of $4k$, where $k$ is a hyper-parameter. From each consecutive $k$ frames, we randomly select one frame and stack them as visual input in training.}
\vspace{-0.3in}
\label{fig:method_demostration}
\end{figure}

We propose to provide RL agents the same transition function in simulation as it has in the real world. We introduce \emph{Multi-Modal Delay Randomization (MMDR)} to provide randomized latency and asynchronous multi-modal observations in simulation. Specifically, we randomize the sampling of the proprioceptive state and visual observation separately to utilize the domain-specific characteristics and simulate independent latency for different modalities.

In simulation for RL, to maintain the fidelity with the real world, the simulation frequency (400 Hz) is several times higher than the control frequency (25 Hz) of the robot. Our method reads the proprioceptive state at every simulation step and uses a queue with a fixed length to store recent historical proprioceptive states. We assume the proprioceptive state is changing smoothly in the real world. Therefore, for each episode, we sample a proprioceptive delay and use linear interpolation to calculate the delayed observation from two adjacent states in the entire buffer with the sampled delay. In this manner, we can provide the same proprioceptive state transition in simulation as in the real-world.

We stack four depth images as visual observations for the RL agent, ideally with one current frame and three historical frames. In this way, the network can capture the robot motion and predict future path plans more effectively. A straightforward way to simulate latency for visual observation is to use the same randomized strategy as for the proprioceptive state. However, due to the on-board computation resource limitation to process the visual data and the mechanism of depth camera, we set depth camera to provide much fewer observations (30 Hz) compared to proprioceptive sensors (around 1K Hz). In this case, the transition between frames is no longer smooth. 
To handle non-smooth visual observation at a low frequency,
we randomize the visual observation in a discrete manner. We obtain the simulated visual observation at every control step and store them in the the visual observation buffer maintaining visual observation in the near past. As illustrated in Figure~\ref{fig:method_demostration}, we store the most recent $4k$ depth maps as our visual observations buffer, split the whole visual observation buffer into four sub-buffers, then sample one depth map from each sub-buffer to create the visual input with randomized latency. 

Though there are multiple kinds of latency as we discussed in Section~\ref{sec:Latency}, for policy trained with RL an in end-to-end manner, what really influences the real-world deployment is the accumulated latency across the different stages. Therefore, the separated randomized latency for proprioceptive state and visual observation is enough for the policy training. Since we randomize the observation from different modalities independently, we provide the same asynchronous multi-modal observation as the proprioceptive sensors and depth camera provide in the real-world.
In this way, we construct the same transition function in the training for the RL policy and make the policy easier to deploy in the real-world.

\noindent\textbf{\emph{{Domain Randomization on Other Parameters.}}}
Besides the randomization for multi-modal latency, we also perform domain randomization on different parameters to bridge the Sim2Real gap. 
The randomized parameters and the corresponding ranges are listed in Table~\ref{table:domain_randomization}. In simulation, the training and testing environments share the same setting. 

Ambient light in the real world and the noisy reading of the depth camera result in missing depth values in the original depth map from the depth camera. During deployment, we fill in all the missing values with the maximum depth (10m). As shown in Figure~\ref{fig:real_visualization}, the shape of obstacles breaks because of missing values, which makes the deployment of the visual-guided locomotion policy even more challenging. To simulate this phenomenon in simulation, we randomly sample 3-30 pixels in each depth map, and set the value of sampled pixels to the maximum depth. Examples of modified depth map are visualized in Figure~\ref{fig:sim_visualization}.

\begin{table}[t]
\centering
\begin{small}
\begin{tabular}{c|c}
\toprule
Parameters & Range  \\
\midrule
Mass ($\times$ default value) & [0.8, 1.2] \\
Motor Friction (Nms / rad) & [0.0, 0.05] \\
Motor Strength ($\times$ default value) & [0.8, 1.2]\\
Lateral Friction (Ns / m) & [0.5, 1.25] \\
Inertia ($\times$ default value) & [0.5, 1.5]\\
Proprioception Latency (s) & [0, 0.04] \\
Kp & [40, 90] \\
Kd & [0.4, 0.8] \\
\bottomrule
\end{tabular}
\vspace{-0.05in}
\caption{Variation of Environment and Robot Parameters.}
\label{table:domain_randomization}
\end{small}
\vspace{-0.2in}
\end{table}

\section{Experimental Results and Analysis}

In our experiments, we evaluate the performance of policies by two metrics: (i) \emph{Moving Distance}: the distance covered by the robot in the forward direction; (ii) \emph{Collision Steps}: the number of time steps where the robot is in contact with obstacles. 
The moving distance represents the ability of an agent to perform the given task while the collision steps reflect the robustness of the policy.

\subsection{Baselines}
\vspace{-0.05in}

\begin{figure}[t]
\centering
\vspace{0.05in}
\includegraphics[width=\linewidth]{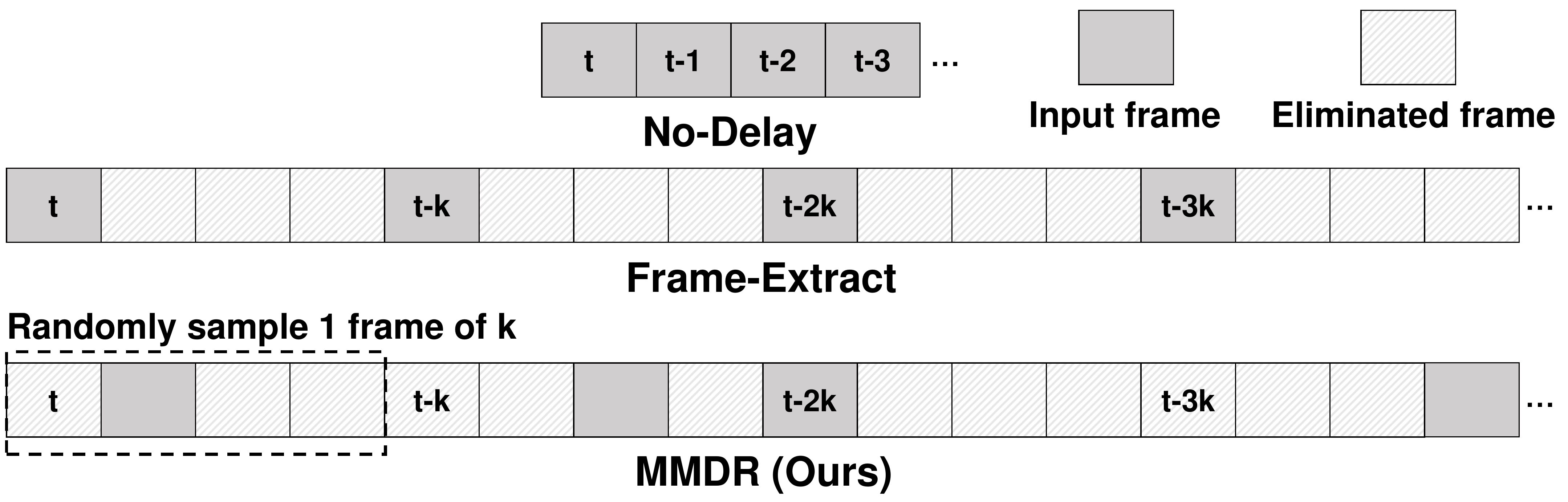}
\vspace{-0.25in}
\caption{\emph{Comparison of Our Method and Two Baselines.} Our method randomly select one frame in each consecutive $k$ frames, while both baselines use the latest image in a fixed interval (1 for No-Delay and $k$ for Frame-Extract).}
\vspace{-0.25in}
\label{fig:comparison_demostration}
\end{figure}

We compare our MMDR with two baselines: 
1) The \emph{No-Delay} baseline is trained without delay randomization, and stacks the most recent $4$ depth maps as visual input. 
2) The \emph{Frame-Extract} baseline is trained without delay randomization, stores the most recent $4k$ frames and stacks the first frames from every consecutive $k$ frames as visual input to provide more temporal information. We empirically found $k=4$ works better.

As shown in Figure~\ref{fig:comparison_demostration}, our method uses the same amount of temporal information as the Frame-Extract baseline, but models the real-world latency and asynchronous multi-modal observation using randomization.

\subsection{Simulation Environment}
\vspace{-0.05in}
\begin{figure}[t]
     \centering
    \vspace{0.05in}
     \begin{subfigure}[b]{\linewidth}
         \centering
         \includegraphics[width=\textwidth]{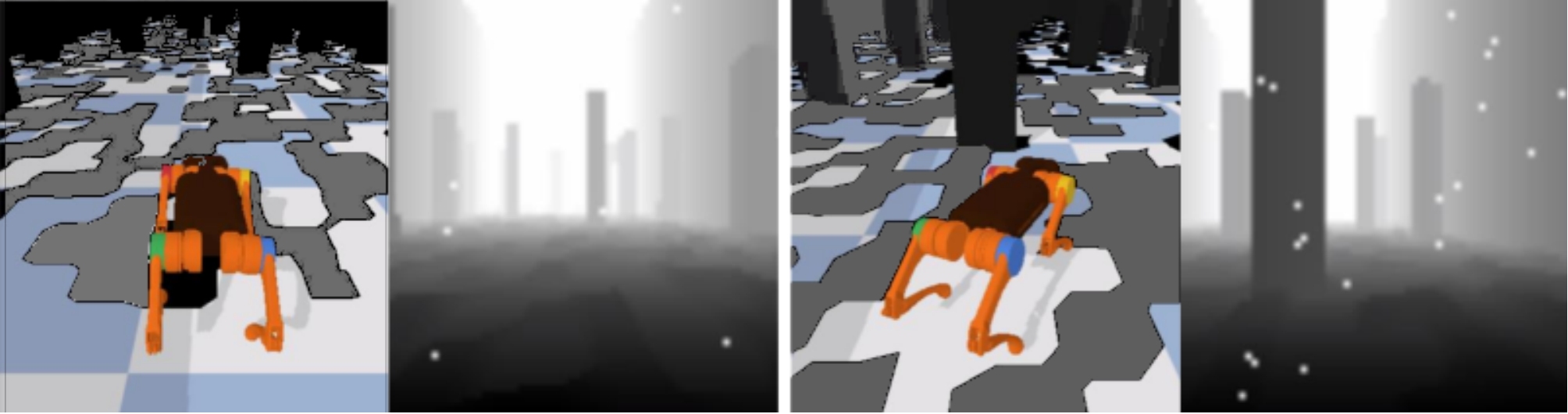}
         \caption{\emph{Simulated Environments.} the environment is shown in the left, and the corresponding observation in the right. Aside from the randomized obstacles and terrain, we add randomly placed white spots in the visual observation to simulate the real observation from the depth camera.}
         \label{fig:sim_visualization}
     \end{subfigure}

     \begin{subfigure}[b]{\linewidth}
         \centering
         \includegraphics[width=\textwidth]{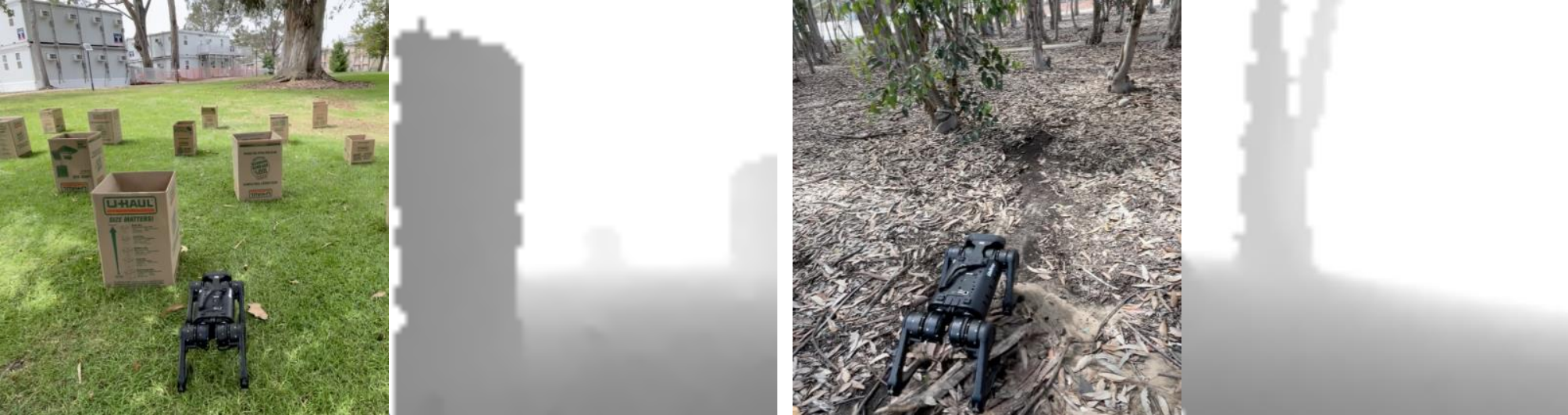}
         \caption{\emph{Real Environments.} We evaluate the generalization ability of policies in real-world scenarios with obstacles of different shapes and sizes on a complex terrain. Due to the noisy reading of the depth camera, some depth values are missing in the depth map, which breaks the shape of an object in the visual observation.
         }
         \label{fig:real_visualization}
     \end{subfigure}
     \vspace{-0.2in}
     \caption{{Samples from the simulator and the real world.}}
     \vspace{-0.1in}
\end{figure}

We perform all our simulation experiments using PyBullet~\cite{coumans2021pybullet}. We train agents to control a Unitree A1\cite{a12021} robot to maneuver in the environment with obstacles and a complex terrain. As shown in Figure~\ref{fig:sim_visualization}, simulated obstacles are cuboid rigid bodies with random positions and shapes, which remain static throughout the episode, and the complex terrain is created with a random height field. 

\subsection{Results in Simulation}
\vspace{-0.05in}

\begin{table}[t]
  \begin{minipage}{0.53\linewidth}
    \centering
    \includegraphics[width=\linewidth]{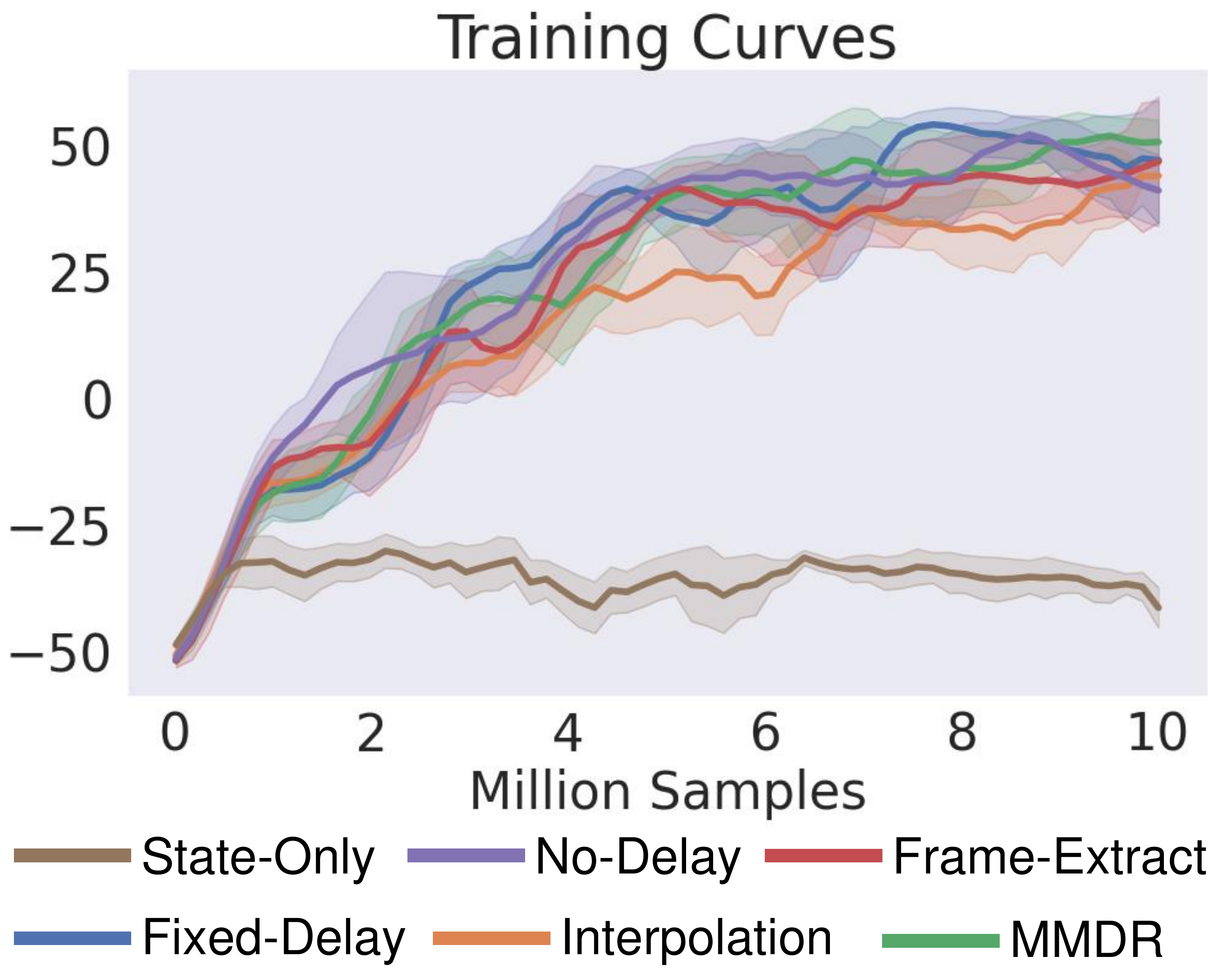}
    \end{minipage}
      \begin{minipage}{0.42\linewidth}

    \begin{tabular}{l|c}
        \toprule
        \multicolumn{1}{l|}{}
            & \multicolumn{1}{c}{\textbf{Moving}}  \\ 
            & \multicolumn{1}{c}{\textbf{Distance~$\uparrow$}}  \\
        \midrule
        State-Only
            & $2.9 \scriptstyle{\pm 0.5}$\\
            No-Delay
             &$ 26.5\scriptstyle{\pm 5.0} $\\
        Frame-Extract
          & $24.9\scriptstyle{\pm 3.1 } $\\
         Fixed-Delayed
            & $18.5 \scriptstyle{\pm 0.8}$\\
         Interpolation
            & $21.4 \scriptstyle{\pm 2.7}$\\
        Ours
            & $\mathbf{28.7} \mathbf{\scriptstyle{\pm 7.7}}$ \\
            \bottomrule
        \end{tabular}
                \label{table:static-sim}
        \end{minipage}
    
    \captionof{figure}{\emph{Training Curves and Evaluation Results.} All methods show similar sample efficiency in the environment that it's trained on, even if MMDR trains in a noisy environment. MMDR yields comparable result to No-Delay during training, and significantly outperforms all baselines in real robot test according to Table V. We did not collect any additional data for representation learning, compared to~\cite{hoeller2021learning}.} 
        \label{fig:training_curves}
            \vspace{-0.25in}
\end{table}

We evaluate all methods by their performance in the environment with random delays varying from 0.04 to 0.12 seconds. Besides the environment with static obstacles, we also evaluate the policies in the environment with moving obstacles to evaluate the generalization ability of learned policies in simulation.

Though MMDR introduces uncertainty into the observation, the MMDR policy show the same sample efficiency and final performance in the training environment as in Figure~\ref{fig:training_curves}. When testing in the same environment with random delays, the policy learned with MMDR moves further with less collisions. This demonstrates the potential to better adapt to the real world. 

To further evaluate the generalization ability of policies, we evaluate all methods in the environment where the obstacles continuously move in random directions at random speeds.
This environment requires agents to have a better ability to understand the dynamic environment.
Table~\ref{table:moving-sim} shows our superiority, in both traversing skill and robustness, in the dynamic environment. MMDR significantly increases the Moving Distance by nearly {\bf 100\%}, and reduces the Collision Steps by {\bf 475\%} to the No-Delay baseline and {\bf 340\%} to  the Frame-Extract baseline. We conjecture that it is because independently randomizing multi-modal observation makes an agent robust to the relative position and speed changes in the environment, which suggest better performance during real-world deployment.

\noindent\textbf{\emph{Comparison Study.}} 
We conduct three sets of comparison studies to show the importance of visual observation for RL training and the random selection mechanism for randomizing visual observation. 

First, we train the policy in the same environment without visual input. We use a 4-Layer MLP for training, corresponding to the \emph{State-Only} baseline in Figure~\ref{fig:training_curves}. The State-Only agent learns almost nothing for such a complex task, which suggests the importance of visual observation for robot to manuever in the complex environment.

Second, we train the \emph{Fixed-Delay} agent in the environment with fixed delay for proprioceptive and visual observation. Though the \emph{Fixed-Delay} agent shows decent training performance, when evaluating in the test environment, the performance of the \emph{Fixed-Delay} agent drops drastically. The performance drop of the \emph{Fixed-Delay} agent shows that only considering the latency in the multi-modal control system is not enough for training robust multi-modal RL policies, and the randomization mechanism for multi-modal observation is necessary.

Third, we train a policy with multi-modal delay randomization but use an interpolation strategy for visual observation randomization other than randomizing the visual observation in a discrete manner. We sample a delay $t_d$ for each episode and use linear interpolation to calculate visual observation from two adjacent visual observations around $t-t_d$ at each step $t$ (corresponding to the \emph{Interpolation} baseline in Figure~\ref{fig:training_curves}). Our MMDR consistently outperforms the \emph{Interpolation} baseline during training and shows better evaluation performance. This result indicates that randomized selection is better for randomizing visual observation. We also find that though the \emph{Interpolation} baseline does not perform well compared to both baselines (\emph{No-Delay} and \emph{Frame-Extract}) during training, when evaluated on an environment with moving obstacles (as shown in Table~\ref{table:moving-sim}), it generalizes better. This suggests that when temporal information in visual observation is crucial for decision-making, it's essential to take latency and asynchronous multi-modal observations into consideration.

\begin{figure}[t]
    \centering
    \includegraphics[width=0.6\linewidth]{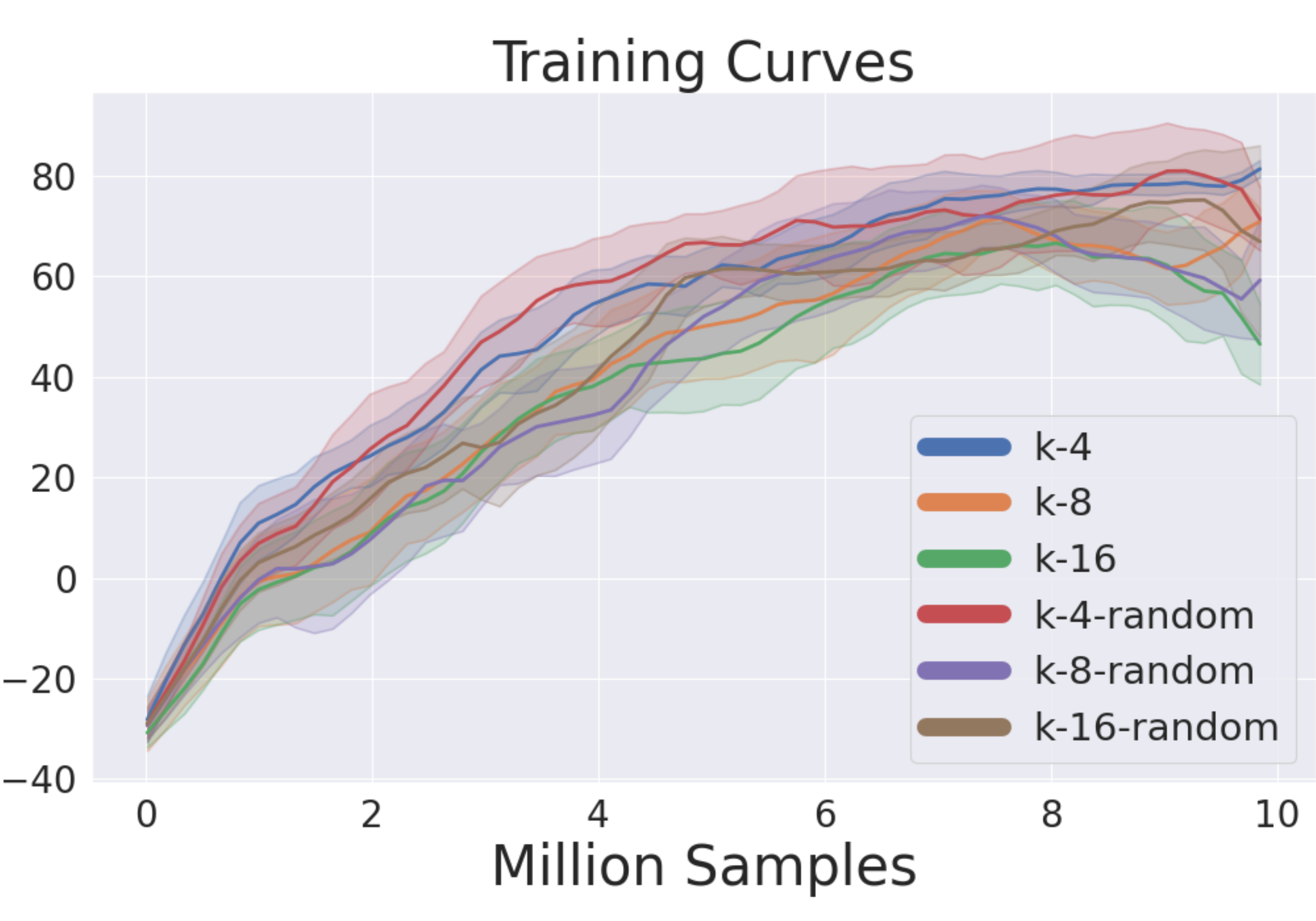}
    \vspace{-0.1in}
    \caption{Ablation Study on $k$. We choose $k=4,8,16$ (the visual input can maintain the information in the past of 0.64s, 1.28s, 2.56s). ``-random'' indicates MMDR is applied.}
    \vspace{-0.15in}
    \label{fig:ablation_k}
\end{figure}

\noindent\textbf{\emph{Ablation on $k$}.} 
We perform ablation on $k$ by and show the training curve in Figure~\ref{fig:ablation_k}. We ablate $k$ on different values (4, 8, 16), which represents information of (0.64s, 1.28s, 2.56s) in the past. We use ``-random'' to indicate MMDR is applied. The policy is generally robust and stable to the change of $k$, and we achieve relatively better results with $k=4$. We conjecture it is important to capture a past history of information and the continuous frames can additionally provide motion information. We find randomizing the frame selection with MMDR does not hurt the training sample efficiency much, but allows much better Sim2Real transfer.

\subsection{Results in the Real World}
\vspace{-0.05in}

\begin{table}[t]
\vspace{0.1in}
    \centering
    \begin{tabular}{l|c|c}
        \toprule
        \multicolumn{1}{l|}{}
            & \multicolumn{1}{c|}{\textbf{Moving Distance~$\uparrow$}} & \multicolumn{1}{c}{\textbf{Collision Steps~$\downarrow$}} \\ 
        \midrule
        State-Only
             &$ 3.5\scriptstyle{\pm 1.3} $ 
            & $ 475.1\scriptstyle{\pm 261.3}$\\
            No-Delay
             &$ 6.0\scriptstyle{\pm 1.7} $ 
            & $401.5\scriptstyle{\pm 118.0}$\\
        Frame-Extract
           &  $6.6\scriptstyle{\pm  2.5} $ 
            &  $287.9 \scriptstyle{\pm 264.5}$ \\
        
        Fixed-Delayed
             &$ 9.9 \scriptstyle{\pm 2.7} $ 
            & $ 545.9 \scriptstyle{\pm 119.9}$\\
            
        Interpolation
             &$ 7.1\scriptstyle{\pm 0.7} $ 
            & $ 94.8\scriptstyle{\pm 12.3}$\\
        Ours
             & $\mathbf{11.4} \mathbf{\scriptstyle{\pm 1.9}}$ 
             &  $\mathbf{84.4} \mathbf{\scriptstyle{\pm 22.9}}$  \\
            \bottomrule
        \end{tabular}
            \captionof{table}{\emph{Generalization evaluation in simulation with moving obstacles.} When evaluated in much more challenging environment with moving obstacles, MMDR not only performs significantly better than baselines in both metrics, but also is much more stable across different seeds.}
            	\label{table:moving-sim}
\vspace{-0.3in}
\end{table}

\begin{table*}
\vspace{0.05in}
\centering
        \begin{tabular}{l|cccc|cccc}
        \toprule
        \multicolumn{1}{l|}{}
            & \multicolumn{4}{c|}{\textbf{Moving Distance~$\uparrow$}} & \multicolumn{4}{c}{\textbf{Collision Count~$\downarrow$}} \\ 
            & Box. & Dense Box. & Moving Box. & Forest. 
            & Box. & Dense Box. & Moving Box. & Forest. \\
        \midrule
            No-Delay
            &$ 444.7\scriptstyle{\pm 115.0} $ 
            & $ 447.6\scriptstyle{\pm 147.6}$ 
            &$ 505.7\scriptstyle{\pm 120.5} $ 
            & $ 733.8 \scriptstyle{\pm 118.0}$
            & $0.0 \scriptstyle{\pm 0.0}$ 
            & $1.0 \scriptstyle{\pm 0.27}$
            & $ 0.33 \scriptstyle{\pm 0.27}$ 
            & $ 0.22 \scriptstyle{\pm 0.16}$\\
        Frame-Extract
            & $358.4 \scriptstyle{\pm 155.3}$ 
            & $280.0 \scriptstyle{\pm 79.8}$ 
             & $ 380.4\scriptstyle{\pm 261.6}$ 
            & $ 572.4 \scriptstyle{\pm 256.7}$ 
            & $0.56 \scriptstyle{\pm 0.42}$ 
            & $1.21 \scriptstyle{\pm 0.47}$
             & $ 0.44 \scriptstyle{\pm 0.62}$ 
            &  $ 0.22 \scriptstyle{\pm 0.32}$ \\
        Ours
             & $\mathbf{859.9} \mathbf{\scriptstyle{\pm 271.4}}$ 
             & $\mathbf{641.2} \mathbf{\scriptstyle{\pm 49.9}}$ 
             & $\mathbf{973.0} \mathbf{\scriptstyle{\pm 148.4}}$ 
             & $\mathbf{992.5} \mathbf{\scriptstyle{\pm 335.0}}$
             & $\mathbf{0.0} \mathbf{\scriptstyle{\pm 0.0}}$ 
             &  $\mathbf{0.9} \mathbf{\scriptstyle{\pm 0.6}}$
             & $\mathbf{0.33} \mathbf{\scriptstyle{\pm 0.0}}$ 
             &  $\mathbf{0.22} \mathbf{\scriptstyle{\pm 0.16}}$  \\
            \bottomrule
        \end{tabular}
        \vspace{-0.05in}
	    \captionof{table}{\emph{Real World Deployment Performance.} MMDR significantly improve  forward efficiency and safety in complex wild environments.}
        \label{table:real}
\vspace{-0.25in}
\end{table*}

\begin{figure}[t]
\centering
\vspace{-0.05in}
\includegraphics[width=\linewidth]{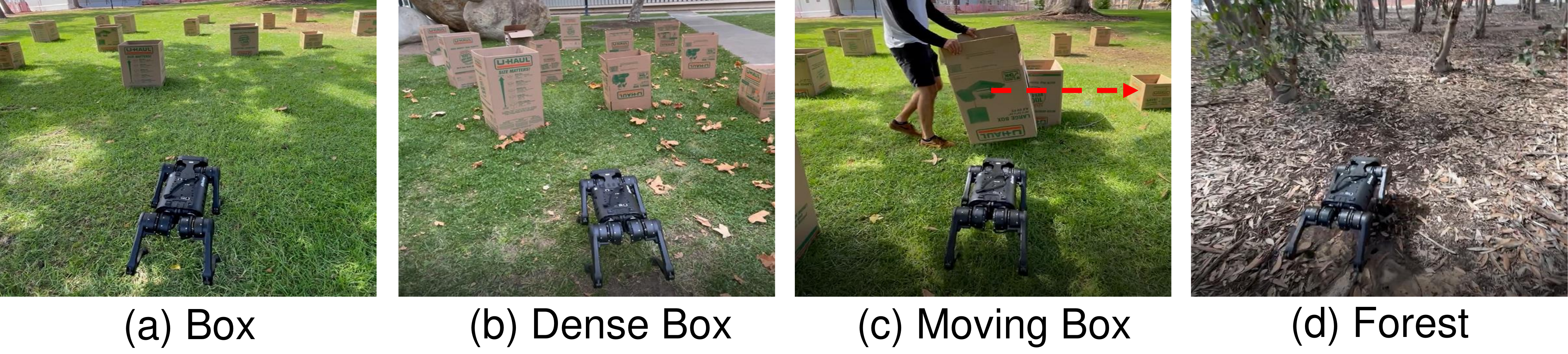}
\vspace{-0.15in}
\caption{Four Real world environments in our experiments.}
\vspace{-0.15in}
\label{fig:real-scenarios}
\end{figure}
Given the performance of different methods in simulation, we conduct real-world experiments with these three methods: Ours, No-Delay, Frame-Extract. We emphasize no human is involved in controlling the robot in the real world, and all computation runs on the onboard TX2. Our policy performs planning and locomotion control on the robot end-to-end.

\noindent\textbf{\emph{Robot Setup.}}
In the real-world, we perform evaluations using the Unitree A1 Robot~\cite{a12021}, a quadruped robot with 18 links and 12 DOF (3 for each leg). In the real-world, we use \textbf{Collision Count}, i.e. number of times the robot hits an obstacle, rather than the Collision Steps used in simulation, since it is dangerous to let the real robot continuously collide with obstacles.
Our policy output actions (target joint angles) at 25 Hz, while the underlying PD controller computes the real torque for the motors at 400 Hz. Realsense updates the depth image buffer in 30 Hz. Kp and Kd are set to 40 and 0.6 respectively.

We first evaluate all the methods in environments similar to the training environment. We place various sized boxes on a sloped lawn with different density. We refer to these environments as \textbf{Sparse box placement (Box.)} (as in Figure~\ref{fig:real-scenarios}(a)) and \textbf{Dense box placement (Dense Box.)} (as in Figure \ref{fig:real-scenarios}(b)). When the robot touches any part of its limbs to an obstacle, such as grazing past a box, that is considered a collision. The episode is terminated when the robot falls over, or if the robot has a head-on collision with an obstacle.

In the \textbf{Sparse box placement (Box.)} and \textbf{Dense box placement (Dense Box.)} environments, as demonstrated in Table~\ref{table:real}, MMDR excels baselines by a large margin. MMDR moves twice as far as the other baselines do without colliding into any boxes in \textbf{Sparse box placement}, and maintains a similar performance when the obstacles become much denser. In contrast, Frame-Extract performs poorly in \textbf{Sparse box placement (Box.)} and the performance becomes worse as the density of boxes increase. We speculate that using visual observations covering a longer time span makes the policy become more sensitive to the latency in the real world and enlarge the asynchrony in the multi-modal observation. This phenomenon suggests that modeling the real world latency and asynchronous multi-modal observations for a learned multi-modal RL policy is essential.

\begin{figure}[t]
\centering
\includegraphics[width=0.9\linewidth]{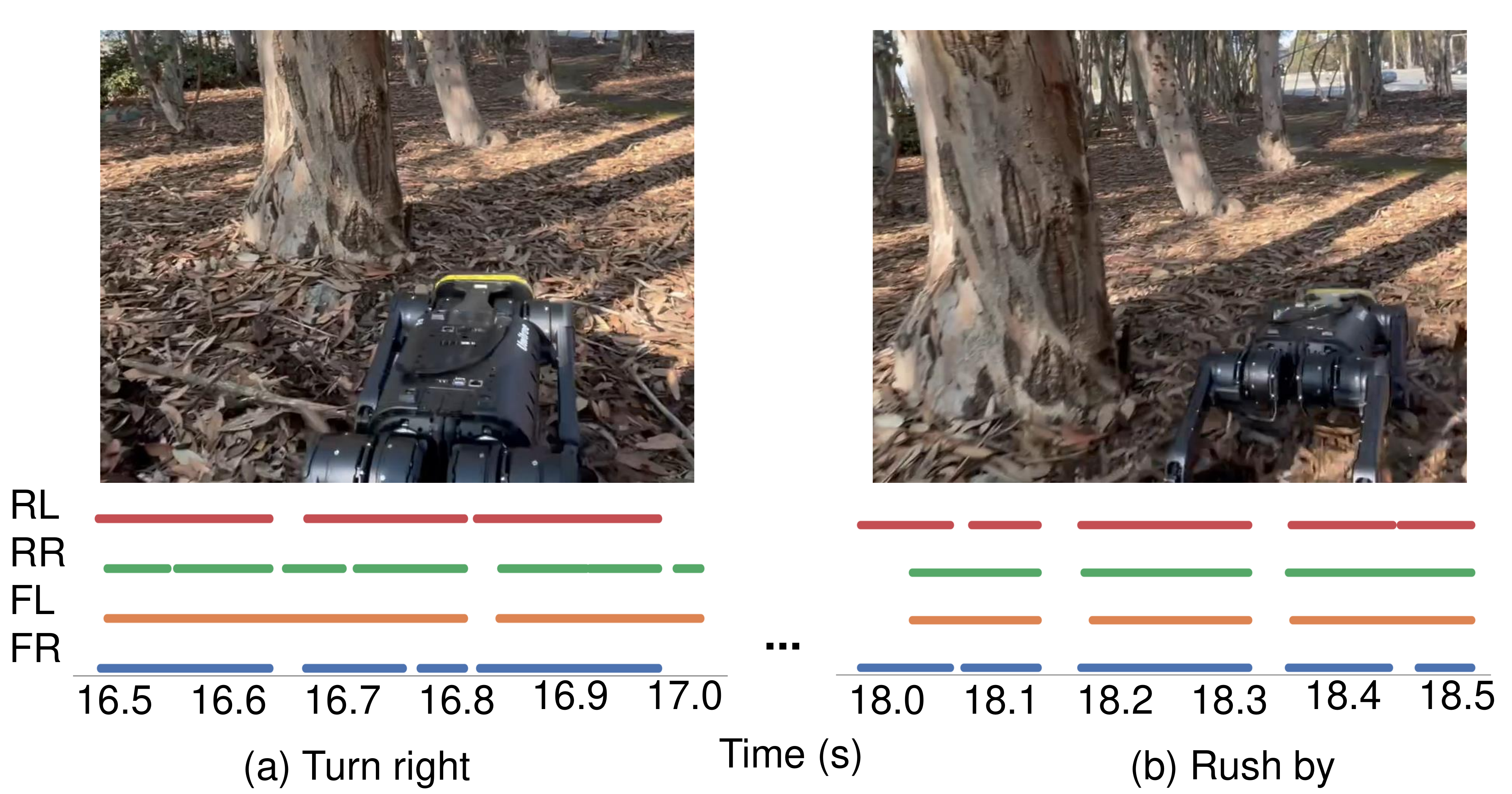}
\caption{\emph{Gait Pattern Analysis.} We plot the gait pattern to analyze locomotion details. Lines denote leg stepping on the ground. During turning right, the front left leg continuously steps to support the twisting while right legs move in higher frequency. After adjusting the direction, robot turns to bouncing gait for higher speed.}
\vspace{-0.2in}
\label{fig:gait}
\end{figure}

Beyond static environments similar to the training environment, we deploy policies in two more challenging scenarios: (i) {\bf Moving box (Moving Box.)}: the box is moving slowly in the same sloped lawn as previous environments shown in Figure~\ref{fig:real-scenarios}(c); (ii) {\bf Forest where the ground is covered by branches and fallen leaves (Forest.)} as in Figure~\ref{fig:real-scenarios}(d). The \textbf{Moving Box} environment is more challenging because of the dynamic obstacles, and the \textbf{Forest} environment is more challenging for the diverse lighting condition, unseen obstacles and complex terrains (the branches, and leaves on the ground). We find that in the {\bf Moving Box} environment, the improvement our method obtained over the baselines further enlarge. 
We conjecture that this happens because when the environment becomes dynamical, the latency in the multi-modal control system becomes more influential. 
If the agent is not able to deal with the latency and asynchronous multi-modal observations, it's more likely to make wrong decisions in the dynamic environment.
This result indicates that it's necessary to simulate latency and asynchronous multi-modal observation in training for learning vision-guided locomotion policy.
In the Forest environment, the policies trained with our MMDR still obtains large improvement over the baselines in the moving distance, while having similar collision results. Stepping on branches or fallen leafs can cause unexpected state transitions that don't happen in the simulation and our policies are more robust to unseen transitions due to our randomization for multi-modal input during training.

To study the learned locomotion skills in more details, we visualize the gaits pattern during testing in Forest environment as shown in Figure~\ref{fig:gait}. We selected two frames when the robot is turning around a tree. In Figure~\ref{fig:gait}(a), the front left leg pushes the ground longer for power and the two right legs take small steps to change the orientation of the body. Once the robot faces to correct direction, it uses the bouncing gait for acceleration to pass the tree (Figure~\ref{fig:gait}(b)). Such a complex locomotion sequence demonstrates that robot combines multi-modal information well to better adapt to the complex environments.
\section{Conclusion}
\vspace{-0.05in}
Latency, as a crucial reality gap, exists in many parts of the robot control pipeline. We propose the Multi-modal Delay Randomization technique to address the latency issues in real-world deployment for vision-guided quadruped locomotion control. Our approach shows great advantages on two realistic metrics both in the simulation and on a real robot. It also improves generalization and adaptation ability in unseen scenarios so as to help the robot pass through arbitrary barriers in the real world. These results suggest MMDR can be universally applied not only to legged locomotion, but potentially to many other visual robotic control tasks as well.

\bibliographystyle{IEEEtranS}
\bibliography{reference}{}

% Appendixes should appear before the acknowledgment.

% \section*{ACKNOWLEDGMENT}
\end{document}